\documentclass{article}
\usepackage{arxiv}

\usepackage[utf8]{inputenc} % allow utf-8 input
\usepackage[T1]{fontenc}    % use 8-bit T1 fonts
\usepackage{hyperref}       % hyperlinks
\usepackage{url}            % simple URL typesetting
\usepackage{booktabs}       % professional-quality tables
\usepackage{amsfonts}       % blackboard math symbols
\usepackage{nicefrac}       % compact symbols for 1/2, etc.
\usepackage{microtype}      % microtypography
\usepackage{lipsum}		% Can be removed after putting your text content
\usepackage{graphicx}
\usepackage{subfigure}
\usepackage{amsmath}

\title{ClsGAN: Selective Attribute Editing Model Based On Classification Adversarial Network}

\author{ \href{https://orcid.org/0000-0000-0000-0000}
		{\includegraphics[scale=0.06]{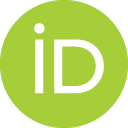}\hspace{1mm}Ying Liu}\\
	College of Informatics\\
	Huazhong Agricultural University\\
	Wuhan, 430070, China 
	\And
	\href{https://orcid.org/0000-0000-0000-0000}{\includegraphics[scale=0.06]{orcid.png}\hspace{1mm}Heng Fan} \\
	Department of Computer Science\\
	Stony Brook University\\
	Stony Brook, 11794, USA \\	
	\And
	\href{https://orcid.org/0000-0000-0000-0000}{\includegraphics[scale=0.06]{orcid.png}\hspace{1mm}Fuchuan Ni} \\
	College of Informatics\\
	Huazhong Agricultural University\\
	Wuhan, 430070, China \\	
	\And
	\href{https://orcid.org/0000-0000-0000-0000}{\includegraphics[scale=0.06]{orcid.png}\hspace{1mm}Jinhai Xiang} \\
	College of Informatics\\
	Huazhong Agricultural University\\
	Wuhan, 430070, China \\	
}

\hypersetup{
pdftitle={A template for the arxiv style},
pdfsubject={q-bio.NC, q-bio.QM},
pdfauthor={David S.~Hippocampus, Elias D.~Striatum},
pdfkeywords={First keyword, Second keyword, More},
}

\begin{document}
\maketitle

\begin{abstract}
	Attribution editing has achieved remarkable progress in recent years owing to the encoder-decoder structure and generative adversarial network (GAN). However, it remains challenging in generating high-quality images with accurate attribute transformation. Attacking these problems, the work proposes a novel selective attribute editing model based on classification adversarial network (referred to as ClsGAN) that shows good balance between attribute transfer accuracy and photo-realistic images. Considering that the editing images are prone to be affected by original attribute due to skip-connection in encoder-decoder structure, an upper convolution residual network (referred to as Tr-resnet) is presented  to selectively extract information from the source image and target label. In addition, to further improve the transfer accuracy of generated images, an attribute adversarial classifier (referred to as Atta-cls) is introduced to guide the generator from the perspective of attribute through learning the defects of attribute transfer images. Experimental results on CelebA demonstrate that our ClsGAN performs favorably against state-of-the-art approaches in image quality and transfer accuracy. Moreover, ablation studies are also designed to verify the great performance of Tr-resnet and Atta-cls.
\end{abstract}

% keywords can be removed

\section{Introduction}
Attribute editing (also termed as attribute transfer) aims to change one or more attributes of images (e.g.,\ hair color, sex, style, etc.) while other attributes remain. The key of attribute editing is to achieve high quality and accurate attribute transfer of generated images. In recent years, generative adversarial network (GAN)\cite{1} has greatly advanced the development of attribute editing. Inspired by this, numerous approaches \cite{11,12,18,19,20,21,22} have been proposed to change local (e.g., hair color, adding accessories, altering facial expressions, etc.) or global (e.g., gender, age, style, etc.) attributes of images.

In addition, in order to obtain accurate attribute transfer images, encoder-decoder structure\cite{24} has been used in attribute editing. Despite promising performance, the method may result in poor quality of generated image because of the bottleneck layer. To address the issue, skip-connection is applied to  encoder-decoder architecture for high image quality\cite{19,20}. Nevertheless, the use of skip-connection brings about the trade-off between image quality and accuracy\cite{20}, i.e., it generates high-quality images at the cost of low attribute accuracy.

Through an in-depth empirical investigation of GAN model\cite{1}, in addition to the original image, the generated image is required to be fed to the discriminator for learning the defects of generated images during the training of discriminator. This way, the discriminator is able to guide the optimization of generator according to the defect information. Besides, for attribute classifier in attribute editing, most recent approaches\cite{18,19,20,21,22} only take as input the original images. Nevertheless, these methods ignore the positive role of generated images on enforcing attribute transfer accuracy when training the classifier. 

\begin{figure*}
	\includegraphics[width=1\textwidth]{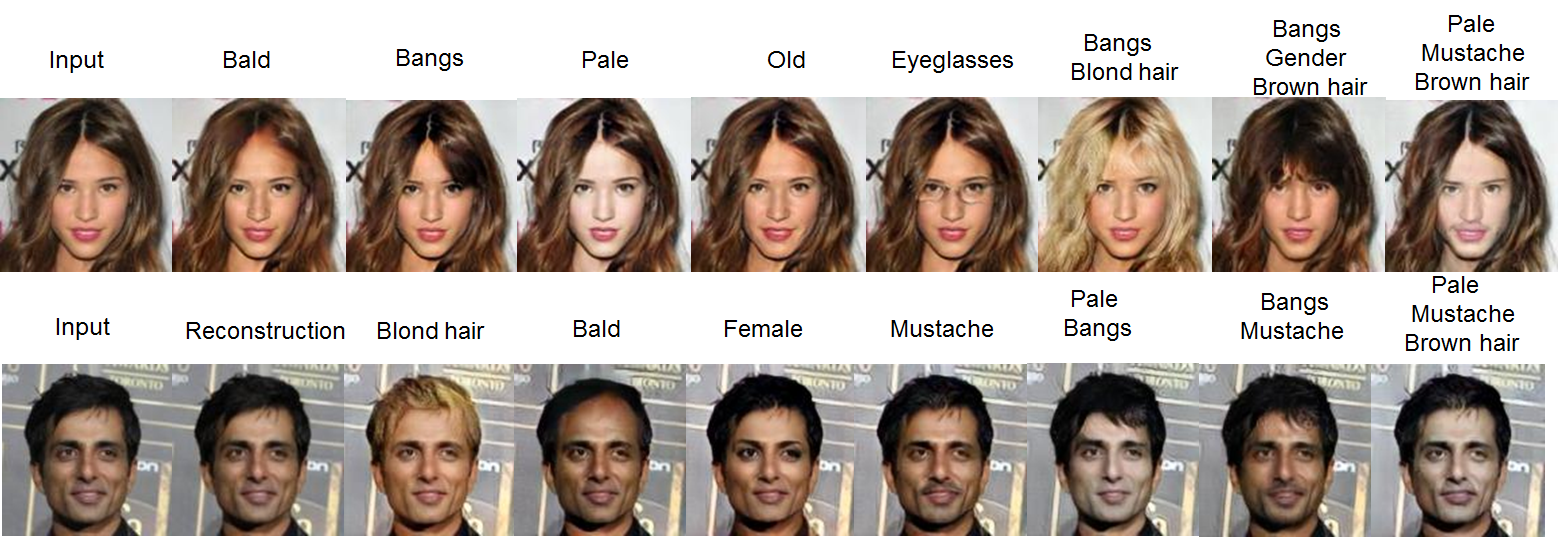}
	\caption{Illustration of generated images with the proposed ClsGAN. These generated images demonstrate high quality and accurate attribute transfer from the visual perspective. Best viewed in color.}
	\label{fig:one}
\end{figure*}

In order to address the aforementioned issues,
a novel selective attribute editing model based on classification adversarial network (referred to ClsGAN for short) is proposed. The key innovation of ClsGAN is an attribute adversarial classifier (referred to Atta-cls for short) that aims at enhancing the classification performance. Sharing similar spirit of GAN model\cite{1}, Atta-cls is implemented as an adversarial network of image attributes. During the training of Atta-cls, both the original and generated images are fed to the classifier in which it specifies that the  attributes of generated image are indistinguishable (similar to the fake property in GAN\cite{1}).

In addition, a detailed empirical analysis on limitation of skip-connect in deep encoder-decoder structure\cite{24} is conducted, and the main reason is caused by transmitting both source attribute and details of images into the decoder. Since more information of source images is encoded, it may decrease the target attribute information, resulting in degraded performance. Motivated by the residual neural network\cite{30},  an effective upper convolution residual network(referred to Tr-resnet for short) which is as a decoder is proposed to get more the target attribute information. Tr-resnet is able to selectively acquire source image and target label information by combining both input and output of the upper convolution residual blocks, leading to more accurate attribute editing and high-quality image generation.

Moreover, drawing on the practices of\cite{21,22}, ClsGAN takes an image as input into two separate encoders (attribute encoder and content encoder) to decouple entanglement between the attribute and unchanged content information. To keep labels continuous, encoded attribute information (i.e., the output of attribute encoder) is also employed to approximating to reference labels. As shown in Fig.\ref{fig:one}, the proposed ClsGAN generates photo-realistic images with accurate transfer attributes visually. 

In summary, the contributions of this work are three-fold:

\begin{itemize}
	\item A novel ClsGAN is proposed, which demonstrates significant improvement in realistic image generation with accurate attribute transfer. In particular, the method presents a simple, yet effective upper convolution residual network(Tr-resnet) to alleviate the limitation of skip-connection in encoder-decoder structure.
	
	\item In order to improve attribute transfer accuracy, an attribute adversarial classifier (Atta-cls) is developed to guide the generator by learning defects of attribute transfer images.
	
	\item Extensive quantitative and qualitative experimental results in face attribute editing demonstrate that the proposed ClsGAN outperforms other state-of-the-art approaches. Furthermore, it is also directly applicable for style manipulation.
\end{itemize}

The rest of this paper is organized as follows. Section 2 discusses the related work of this paper. Section 3 illustrates the proposed approach in detail. Experimental results are demonstrated in Section 4, followed by the conclusion in Section 5.

\section{Related works}

Generative adversarial networks (GANs) \cite{1} are defined as a minimax game with a generator and a discriminator in which the generator generates images as photo-realistic as possible and the discriminator tries to distinguish the synthetic images from the original images. Since then, various GANs and GAN-like variants are proposed to enforce the quality of image or stability of training, including designing novel generator/discriminator architecture\cite{2,8,13}, the choice of loss function~\cite{14}, the study of regularization techniques \cite{3,4,15}. What's more,\cite{16,17} introduce an encoder-decoder structure to obtain images' higher-level semantic information and render reconstruct images. VAE/GAN\cite{13} combines VAE\cite{17} with GAN\cite{1} to modify latent expressions of images by reconstructing loss and adversarial loss. GANs have been applied to various fields of the computer vision, e.g.,\ image generation\cite{1,2,9}, image transfer\cite{6,7}, super-resolution image\cite{10}, image deblurring \cite{27}. 

Meanwhile, CGAN\cite{5} takes the reference label as inputs of generator and discriminator to produce specific images that are consistent with the label. Inspired by CGAN, the community makes a large number of contributions in style transfer\cite{6,7} and attribute editing\cite{Li,12}. About style transfer, Pix2pix\cite{6} and CycleGAN\cite{7} realize mutual transformation between two domains about paired and unpaired data respectively. There are also some double domains' transformation models\cite{Li,12} in attributing editing. However, the number of models increases exponentially with the increase of domains by double domains' transformation method, which is not universal and leads to model overfitting and poor generalization ability.	

To address the issue, recent methods mostly employed a classifier to realize attribute classifiaction and transformation. StarGAN\cite{18} takes domain classification restriction to control the attribute transformation of images, along with reconstruction loss, adversarial loss and classification loss. Notably, \cite{19,20,29} apply the skip-connection or its variants with the encoder-decoder structure to render photo-realistic images. To avoid the effects of irrelevant attributes, on the one hand, STGAN\cite{20} and RelGAN \cite{25} both take difference attribute labels as the input. On the other, AME-GAN\cite{21} and AGUIT\cite{22} both separate the input images into image attribute part and image background part on manifolds to avoid entanglement. In this work, ClsGAN is presented, which applies Tr-resnet and attribute adversarial classifier to improve image quality and attribute transfer accuracy.	

\begin{figure*}
	\centering
	\includegraphics[width=0.8\linewidth]{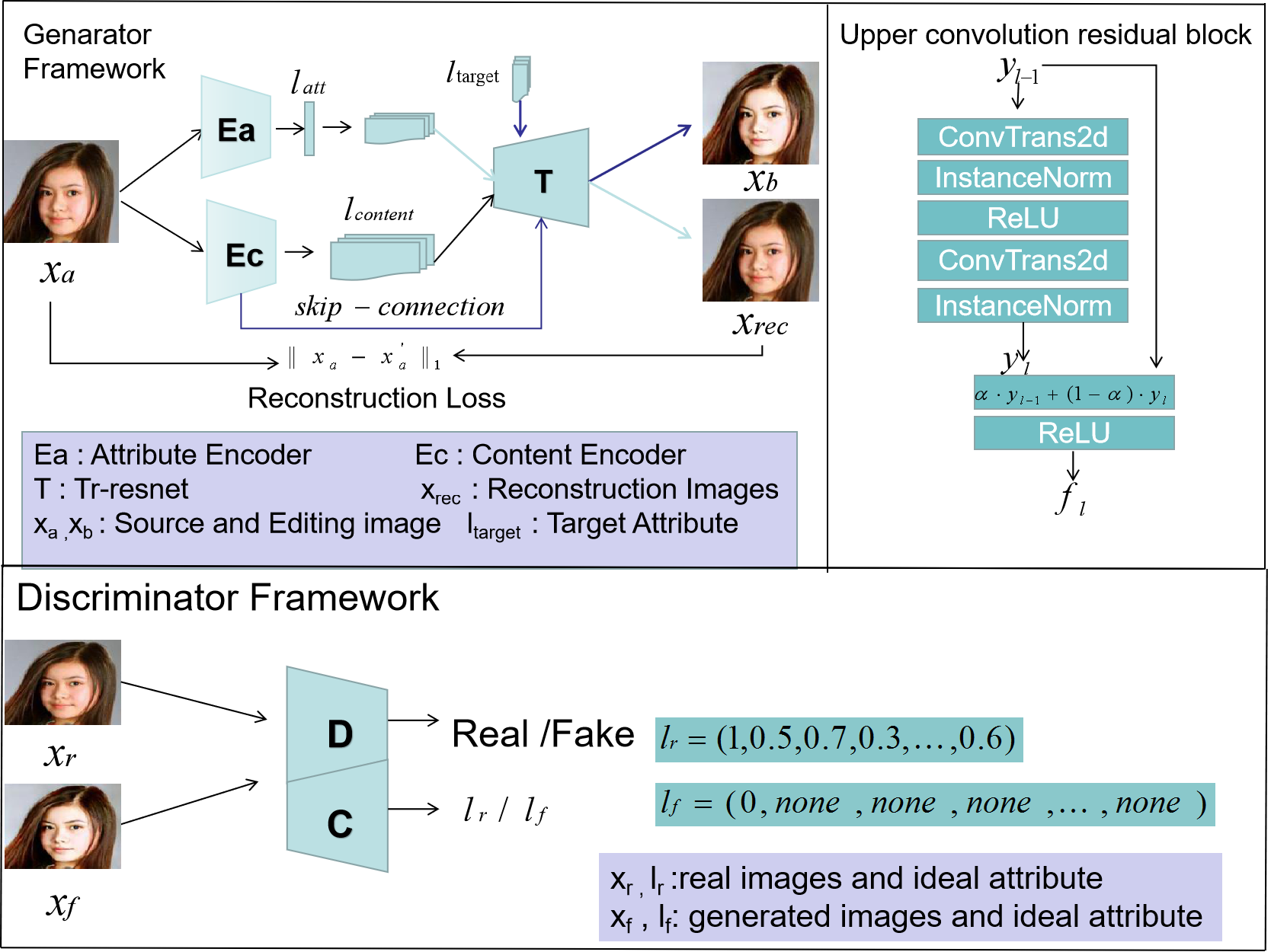}
	\caption{The structure of ClsGAN, which mainly includes the framework of generator and discriminator. The generator is composed of two encoders and a Tr-resnet(T), which consist of a series of convolution layer and upper convolution residual block(upper right) respectively. Discriminator is composed of classifier C (which is the attribute discriminator of Atta-cls) and adversarial discriminator D, whose parameters are shared.}
	\label{fig:two}
\end{figure*}

\section{The Proposed Method}

In this section, ClsGAN for arbitrary attribute editing is described in detail. Section 3.1 introduces the proposed upper convolution residual network (Tr-resnet). Section 3.2 illustrates attribute continuity processing. After that, an attribute adversarial classifier (Att-cls) is employed to enhance attribute transformation accuracy in Section 3.3. The overall network structure and loss function of our ClsGAN are presented in Section 3.4 and Section 3.5.

%-------------------------------------------------------------------------
\subsection{Upper convolution residual network (Tr-resnet)}

The skip-connection\cite{19} has been proven to be beneficial to improve the quality of generated image. However, this improvement is obtained at the sacrifice of attribute classification performance. In order to solve this problem, the work of\cite{19} introduces selective transfer units as a novel skip-connection structure(referred to as STU in short). Nevertheless, the STU requires more parameters and computation resource, which severely limits the application. 

% STGAN~\cite{20} proves that skip-connection in AttGAN~\cite{19} is beneficial to improvement of image quality at the cost of attribute classification accuracy, so STGAN constructs a new skip-connection structure called STU. However, it requires more parameters than AttGAN, and the process is relatively complex. 

In this paper, by empirically investigating skip-connection, the reason causing limitation is that the incorporation of source attribute information weakens the target attribute information in decoder. Motivated by the residual structure in\cite{20}, the upper convolution residual network (Tr-resnet) is propoded. Through using Tr-resnet block as a basic unit, a simple yet effective decoder is developed. As shown in the top right of Fig.\ref{fig:two}, each Tr-resnet block takes the combination of the layer's input and output as the unit's output. Furthermore, to effectively make use of resource image and target attribute information, Tr-resnet applys the weighting strategy to resource image information from encoder, input and output information of each Tr-resnet block in a special unit. Mathematically, the Tr-resnet is represented as follows,

\begin{equation}
y_l^* = \mathrm{Transpose}(y_l)
\label{eq:commutative}
\end{equation}
\begin{equation}
f_{l}=\left\{
\begin{aligned}
&{\alpha\cdot y_{l-1}^*+(1-\alpha)\cdot y_{l}} &(l \in\{1,2,4,5,6\}) \\
&{\alpha\cdot y_{l-1}^*+(1-\alpha)\cdot y_{l}+\beta\cdot x_2  }&(l=3) 
\end{aligned}
\right.
\label{eq:commutative}
\end{equation}

where $y_l$ denotes the Tr-resnet feature of the $l$-th layer, $\mathrm{Transpose(\cdot)}$ represents transposed convolution operation that matches the solution between input $y_l$ and output $y_{l+1}$. $x_2$ denotes the encoder feature of the $2$-nd layer, $f_l$ denotes the output of $l$-th Tr-resnet block. In equation (2), when $l=3$, the Tr-resnet block takes the weighted sum of the 2-nd layer feature map information of the encoder, the 3-th layer input and output information of Tr-resnet as output. When $l\ne3$, the output of Tr-resnet block is only the incorporation of information about the input and output of $l$-th layer. The model initializes $\alpha=(a_1,a_2,...a_s)$, $\beta=(b_1,b_2,..b_s)$, where $a_i,b_i\sim uniform(0,1)$ and $s$ is the number of feature map in $y_{l}$ or $x_2$.

%-------------------------------------------------------------------------

%-------------------------------------------------------------------------
\subsection{Attribute continuity processing}

The approaches in\cite{18,20} are able to generate dual-domain (0 or 1) transfer images. However, the methods are difficult to render various variants with a same attribute, and attribute continuity cannot be guaranteed. To solve this issue, the work of\cite{19} employs a style controller to realize multi-modal transformation for a specific attribute on the basis of the source model, achieving good performance. Motivated by this, the similar is utilized to control the attribute continuity in this work. In specific, the attribute value is obtained by approximating encoded attribute label to the target attribute label. In detail, the optimization object is formulated as follows,

\begin{equation}
{L_{a}} = ||l_r - {E_{a}}({x_r})|{|_1}
\label{eq:commutative}
\end{equation}
where $x_r$ and $l_r$ denote source image and reference label, respectively. $E_a$ represents attribute encoder in generator, which is a general convlution
neural network and takes Convolution-InstanceNorm-ReLU as a unit. The output of $E_a$ has same size with reference label $l_r$. $\|\cdot\|_l$ denotes $l_1$ loss.
%-------------------------------------------------------------------------

\subsection{Attribute adversarial classifier (Atta-cls)}

For attribute classifier, most existing approaches\cite{18,19,20} only take the source image as input and then exploit the optimized classifier to improve the generator. However, it is difficult for these approaches to discover the attribute difference between the generated images and source images. Inspired by the GAN model\cite{1} that optimizes the generator according to the deficiency of generated image learned from the discriminator, an attribute adversarial classifier (Atta-cls) is proposed based on the adversarial method.

In our model, the attribute classifier is designed as an adversarial network. The source image and the generated image are both fed to optimize the classifier, and then the generator is trained according to the defects of generated image.  

Specifically, the target of classifier first evaluates all attribute's distinguishability (similar to the fake/real nature in GAN), and then focuses on the single attributes. When the input is source images, ideally the category is distinguishable (the value is defined as 1 or true) and at the same time the single attribute value should be consistent with label. So the classifier needs to optimize the whole attribute and all single attribute for the source images. In contrast, classifier only needs to assume that it is inseparable for generated images (the value is 0 or false), so the remaining single attributes are  considered needlessly. The detailed operation is shown in Fig \ref{fig:two} (bottom row). Meanwhile, in order to maintain the stability of the model, ClsGAN adds a penalty function for classification loss. The concrete operation is implemented by the loss function. The loss functions of attribute adversarial net(about generator and classifier) are as follows:
\begin{equation}
\begin{aligned}
{loss_{c}} = E_{x_r\sim P_{data}}\{t_r^T\log(C(x_r))\}+E_{x_f\sim P_g}\{(1-t_f^{(1)})\log(1-C(x_f)^{(1)})\}
\label{eq:commutative}
\end{aligned}
\end{equation}

\begin{equation}
\begin{aligned}
\mathop {L_C{_d}} = {-loss_{c}} + \lambda {E_{{x^*}}}[{(||\nabla {x^*}C({x^*})|{|_2} - 1)^2}] 
\label{eq:commutative}
\end{aligned}
\end{equation}

\begin{equation}
\begin{aligned}
\mathop {L_C{_g}} = E_{x_r\sim P_{data},G(x_r,l_f)\sim  P_g}\{t_f^T\log(C(G(x_r,l_f)))\}
\label{eq:commutative}
\end{aligned}
\end{equation}
where $\lambda$ is the scale of gradient penalty and sets to 30. $L_C{_d}$ and $L_C{_g}$ denote the loss functions when training classifier($C$) and generator about attributes. $(x_r,t_r)$ and $(x_f,t_f)$ correspond to the image and label of source domain $P_{data}$ and generation domain $P_g$ respectively, where $t_r/t_f\in R^{n+1}$. Specially, the first element of $t_r/t_f$ is used to evaluate whether the whole attribute is distinguishable or not, define $t_r^{(1)}=1,t_f^{(1)}=0$, and the remaining n-dimensions vector represents image's attributes label(source attribute label $l_r$ and target attribute label $l_f$). $C(x_f)^{(1)}$ denotes the first element in vector $C(x_f)$, $T$ stands for transpose operation. $E_{{x^*}}$ denotes gradient penalty term about $x^*$ which is obtained by line sampling between the original and the generated images.

\subsection{Network structure}

Fig. \ref{fig:two} shows the framework of ClsGAN, in which the generator is comprised of encoders and Tr-resnet. The encoders consist of two convolutional neural networks $E_c,E_a$, whose targets are to extract image contents and attributes information respectively. $E_c$ obtains $512\times16\times16$ high-level semantic content features about source image  and $E_a$ evaluates the attribute label as the basis of label continuity operation.

The Tr-resnet concatenates the content feature from $E_c$ and difference attribute label $l^*=l_f-E_a(x_r)$(it is resized to the same resolution as the content feature) to construct a new feature vector. Then Tr-resnet takes the new feature vector as input to generate reconstructed images or images with specific attributes. For the purpose of selective use of attribute information and original image information, Tr-resnet incorporates residual structure into upper convolutional layers and constructs the Tr-resnet block. The Tr-resnet block structure is shown in Fig. \ref{fig:two} (top right).

The discriminator $D$ consists of a series of convolution layers, and it shares parameters with the classifier $C$ (except for the last layer). The source image and generated image are both used as the input of discriminator and classifier. It is assumed that the image attribute label is $n$ dimension. And the output vector of classifier is $n+1$ dimension. The first dimension is used to distinguish whether the attribute is separable or not and the remaining $n$ dimension vector corresponds to the n-dimensional attributes of the images. By referring to the method of loss function in target detection\cite{31}, the classification vector of the generated image only takes the first dimension for loss function operation, and the other dimensions are expressed as none in the training classifier stage. The specific method is shown in Fig \ref{fig:two} (bottom row).

%-------------------------------------------------------------------------
\subsection{Loss function}
{\bf Adversarial loss}
Similar other models, the work uses GAN to ensure the generated images fine result in quality.
In order to stabilize training, our adversarial loss adopts WGAN-GP~\cite{4}.
\begin{equation}
\begin{aligned}
\mathop  {L_{{D}}} = -({E_{x_r\sim P_{data}}}{D}(x_r)-{E_{x_f\sim P_g}}{D}(x_f)) +\lambda {E_{x'}}[{(||{\nabla _{x'}}{D}(x')|{|_2} - 1)^2}]
\label{eq:commutative}
\end{aligned}
\end{equation}

\begin{equation}
\begin{aligned}
\mathop {L_{{G}}} = {E_{x_r\sim P_{data},G(x_r,l_f)\sim P_g}}{D}(G(x_r,l_f))
\label{eq:commutative}
\end{aligned}
\end{equation}
where $x'$ is obtained by the linear sampling between the original image and the generated image and $\lambda$ is the scale of gradient penalty and sets to 10. $L_D$ and $L_G$ respectively represent the general adversarial loss about discriminator($D$) and generator($G$). 
Generator(G) is composed of encoder $E_c,E_a$ (representing content encoder and attribute encoder respectively) and Tr-resnet ($T$). The relationship between $G$ and $E_c,T,E_a$ is as follows:

\begin{equation}
\begin{aligned}
G(x_r,l_f) = {T}({E_c}(x_r),l_f-E_a(x_r))
\label{eq:commutative}
\end{aligned}
\end{equation}

{\bf Reconstitution loss}
StarGAN reconstructs the original images by means of cycle consistency loss, which will increase the lack of image generation during the cycle. In contrast, ClsGAN uses the attribute difference label vector $l=l_r-E_a(x_r)$ and then directly takes the label and content features into the Tr-resnet to reconstruct the image. The reconstruction loss function is as follows:

\begin{equation}
\begin{aligned}
{L_{rec}} = ||x_r - {T}({E_c}(x_r),l)|{|_1}
\label{eq:commutative}
\end{aligned}
\end{equation}
where the $L_1$ norm is used to suppress blurring of reconstitution images and maintain clarity.

{\bf Object model}
Considering formula (5),(7), the target loss functions of joint training discriminator D and classifier C can be expressed as:

\begin{equation}
\begin{aligned}
\mathop {\min }\limits_{CD} {L_{CD}} =  {\lambda _0}{L_{{D}}} + {\lambda _1}{L_{{C_d}}}
\label{eq:commutative}
\end{aligned}
\end{equation}

The objective function of the generator is comprised of adversarial loss $L_G$, attribute classification  adversarial loss $L_{C{_g}}$, reconstruction loss $L_{rec}$ and attribute continuity loss $L_a$:

\begin{equation}
\begin{aligned}
\mathop {\min }\limits_G {L_{all}} =  - {L_{{G}}} - {\lambda _2}{L_{C{_g}}} + {\lambda _3}{L_{rec}} + {\lambda _4}{L_{a}}
\label{eq:commutative}
\end{aligned}
\end{equation}
Where $L_{{C_d}} ,L_{C{_g}}$ denote attribute classification adversarial losses of classifier and generator, which is mentioned in section 3.2. $\lambda _0,\lambda _1 ,\lambda _2 ,\lambda _3$  and $\lambda _4$  are model tradeoff parameters.

%------------------------------------------------------------------------

\section{Experiments}

Adam optimizer is adapted to train the model, and its' parameters are set to $\beta_1=0.5,\beta_2=0.999$. The learning rate of the first $10$ epoch is set as $2\times 10^{-4}$ , and it is linearly attenuated to $0$ at the next $10$ epoch. In all experiments, the parameters are  $\lambda_0=4,\lambda_1=3,\lambda _2 =1,\lambda _3=20$  and $\lambda _4=1$. All experiments are both performed in a Pytorch environment, with training on a single NVIDIA TESLA V100. Source code can be found at \url{https://github.com/summar6/ClsGAN}.

\subsection{Facial attribute transfer}
\textbf{Dataset} This work adopts CelebA\cite{32} dataset for training and testing of facial attribute editing. The CelebA dataset is a large face dataset, which contains more than $200,000$ images of celebrities' faces and $40$ facial attributes. In this paper, the last $2000$ images of the dataset are set as the test set, and the remaining images are all used as the training set. The model first performs center cropping the initial size $178\times218$ to $170\times170$, then resizes them as $128\times128$ for training and test images.

$13$ of the $40$ attributes are selected for attribute transfer in the paper, which are "Bald", "Bangs", "Black Hair", "Blond Hair", "Brown Hair", "Bushy Eyebrows", "Eyeglasses", "Gender", "Mouth Open", "Mustache", "No Beard", "Pale Skin" and "age". These attributes already covers the most prominent of all attributes.

\begin{figure*}
	\centering
	\subfigure[local editing]{
		\includegraphics[width=1\textwidth]{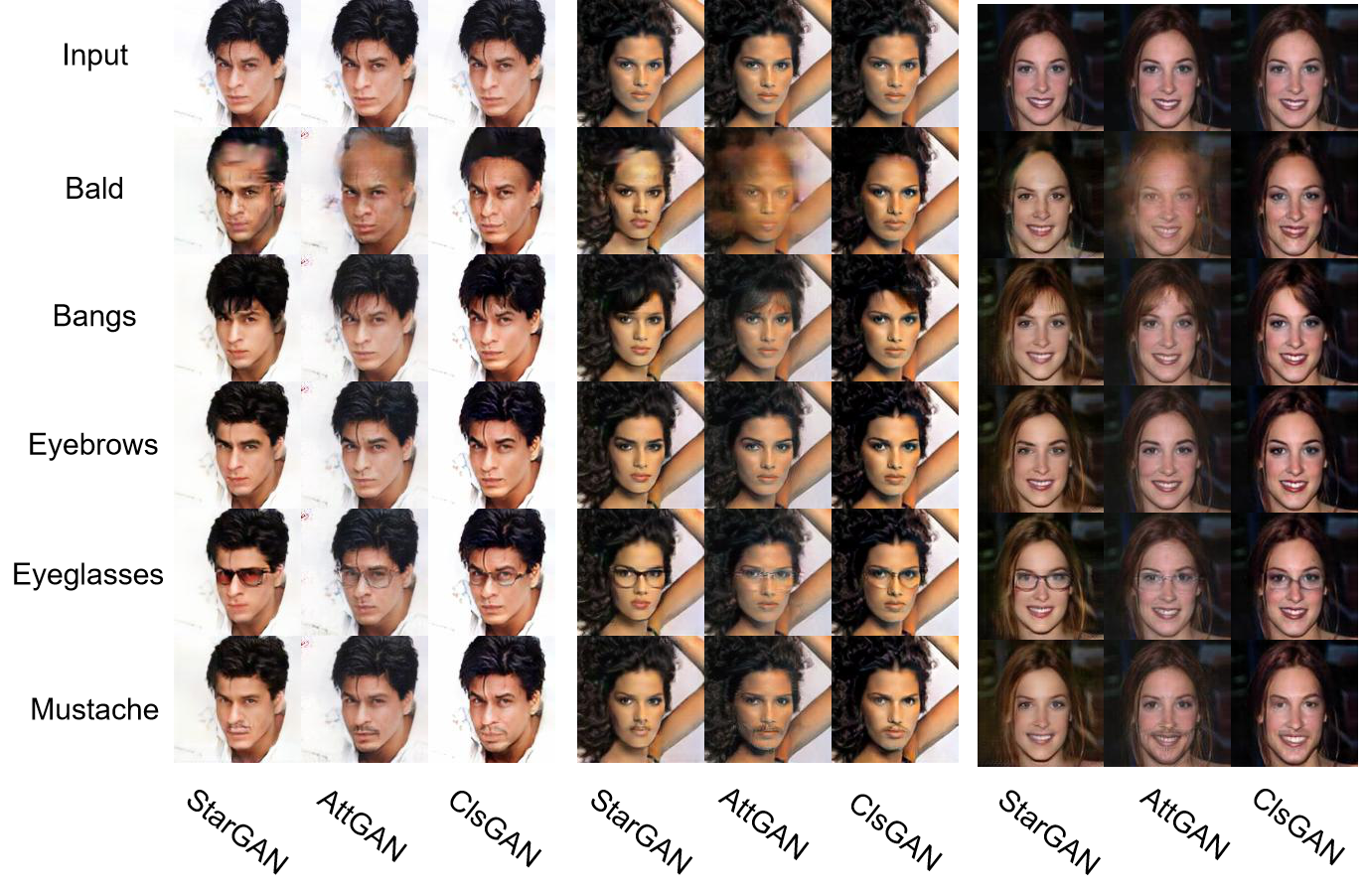}}
	\label{fig:four}
	\hspace{0.01in}
	\subfigure[global editing]{
		\includegraphics[width=1\textwidth]{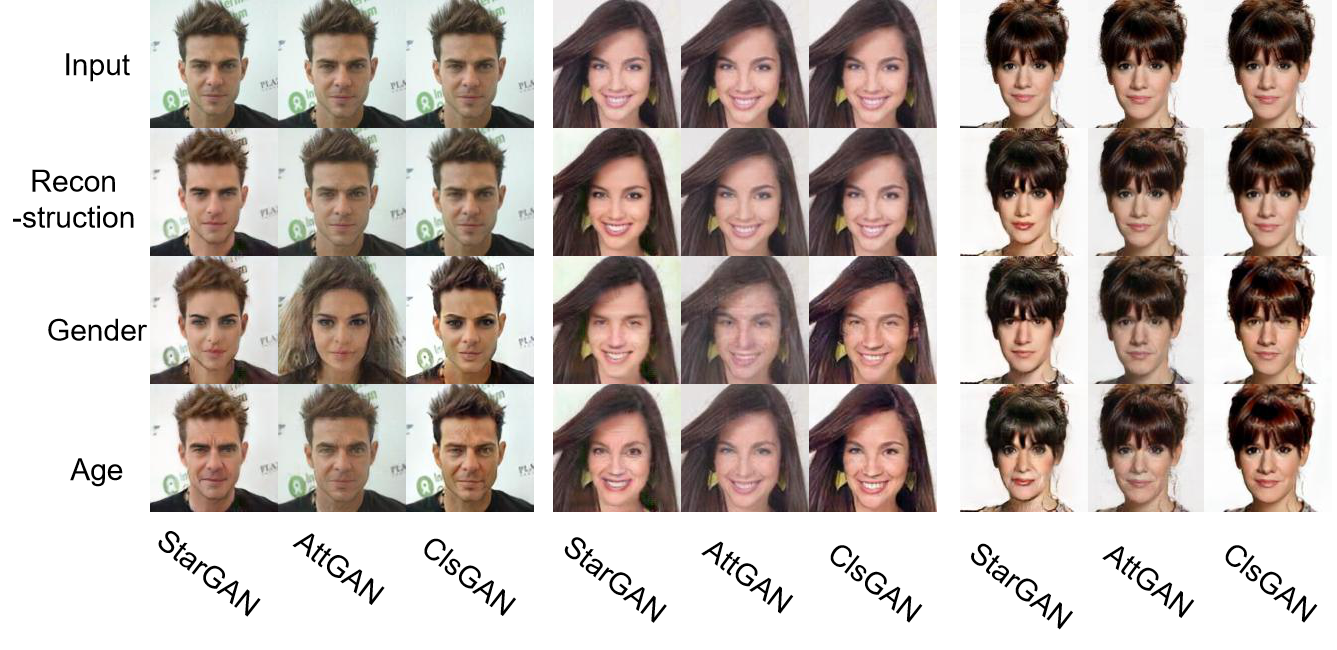}}
	\caption{Face transfer images on CelebA dataset between StarGAN, AttGAN and ClsGAN.}
	\label{fig:three}	
\end{figure*}

\begin{figure*}
	\centering
	\includegraphics[width=1\textwidth]{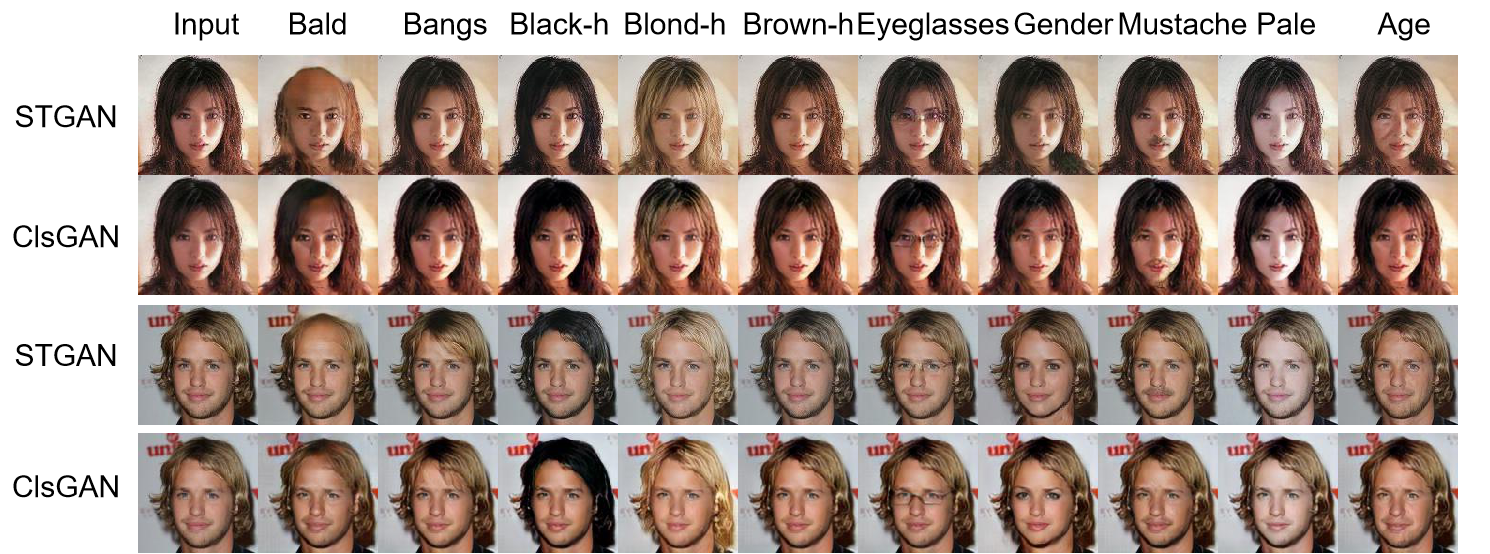}
	\caption{Face transfer images on CelebA dataset between STGAN and ClsGAN.}
	\label{fig:five}
\end{figure*}

\begin{figure}
	\begin{center}
		\includegraphics[width=0.6\textwidth]{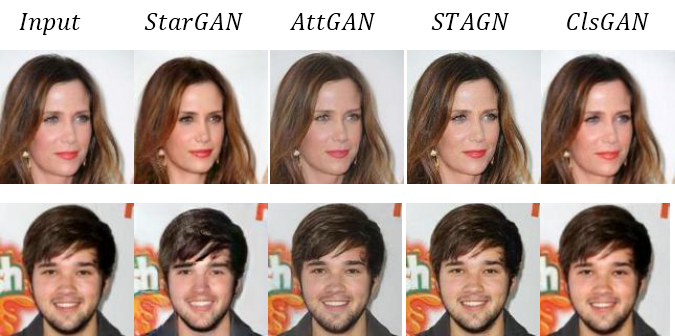}
	\end{center}
	\caption{Face reconstructive images on CelebA dataset between different models.}
	\label{fig:six}
\end{figure}

\begin{figure*}
	\begin{center}
		\includegraphics[width=1\textwidth]{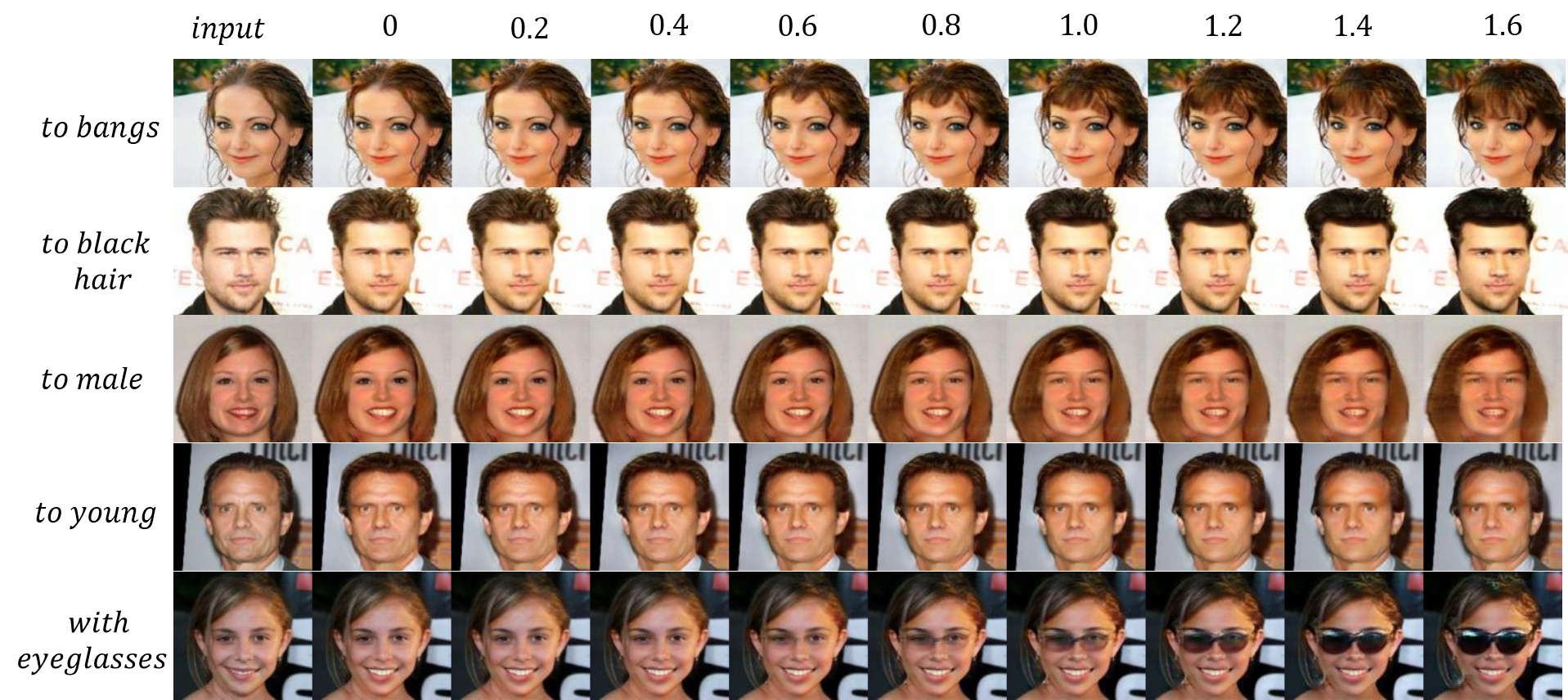}
	\end{center}
	\caption{Interpolation results for facial attributes on CelebA dataset by employing our model. Values among $0$-$1.6$ are the label values about the attribute.}
	\label{fig:seven}
\end{figure*} 

\textbf{Qualitative Assessment}
The work compares the proposed ClsGAN with StarGAN, AttGAN, STGAN in terms of performance of facial attribute transfer. As can be seen in Fig \ref{fig:three} (a), while StarGAN firstly achieves the multi-attribute editing using a single model, it is still limited in manipulating large range attribute. For example, there are obvious blurs and artifacts when editing Bald and Bangs attributes, which makes the images look unrealistic. This may be because it is difficult to take full advantage of the attribute information and other content information by only convolution-residual structure. AttGAN performs better on the attribute editing and the facticity, but the results  contain some differences in background compared with the original images while ClsGAN has a higher degree of restoration in the aspects of background color and skin color(see Fig \ref{fig:three} (b)). In addition, when
performing large range editing or additive attribute, e.g., Bangs, Gender, Eyeglasses and Mustache, there are some blurs and artifacts(see Fig \ref{fig:three}).
One possible reason is that the model still lacks strong implementation capability about attribute information only using the skip-connection technique of encoder-decoder and classification loss. Compared with StarGAN and AttGAN, ClsGAN accurately edits all of the attributes(global and local attributes), which credits to the applying of attribute adversarial classifier. It can be observed from Fig \ref{fig:three}, our results also look more normal and realistic, which benefits from Tr-resnet. 

As can be seen in Fig \ref{fig:five}, both results of STGAN and ClsGAN can accurately edit the attribute and generate more realistic images, however, STGAN is still likely to be insufficiently modified and shows blurred images (when editing attribute Bald) while ClsGAN has a more excellent performance. About reconstructive performance
(Fig \ref{fig:six}), the images show more consistent with original images whatever content, background and other aspects compared with competitive models.

As for attribute continuity, the work not only tests the synthetic images with binary attribute, i.e., with(1) or without(0), but also assumes the attribute value is continuous and may be bigger than 1. As can be seen in Fig \ref{fig:seven} , the transfer images about different values of a single attribute are all presented to analyze the effects of attribute approximation method, where lists the editing images of ClsGAN with attribute label of 0, 0.2, 0.4, 0.6, 0.8, 1, 1.2, 1.4, 1.6 respectively. It can observe that the performance of the attribute gradually increases with the value large, which indicates that attributes have continuity. The transformation effects and image facticity
are excellent from a visual point of view.

\textbf{Quantitative evaluation}

The performance of generated images mainly needs to focus on three aspects, i.e., image quality, reconstruction accuracy and transfer accuracy. ClsGAN's purpose is to maintain the balance between  quality and accuracy. The images generated by the competitive methods recently are either of low quality and high accuracy, or of low accuracy and high quality. Comparing to StarGAN(FID 7.9), ClsGAN greatly improves the image quality with the FID 6.09 while maintaining relatively high conversion efficiency(average 0.66). The method enhances the attribute accuracy comparing with AttGAN(average 0.637) and STGAN(average 0.59) and the image quality is almost equal with STGAN. Meanwhile, our model effectively yields transfer images to some special attributes, which is not easy to convert in other methods, such as Mustache(0.523). The result benefits from the Atta-cls.

ClsGAN utilizes two metrics, FID and SSIM, to evaluate the image quality and the similarity between reconstructed and original images, respectively. The FIDs of ClsGAN and the competitive methods are all shown in Table \ref{Tab1},
where the test dataset of ClsGAN is adapted as the input to randomly edit $10,000$ images for evaluating image quality. Our method is superior to StarGAN, AttGAN and STGAN in image quality with the FID 6.09, in which the result benefits from the application of Tr-resnet. Furthermore, the reconstruction rate outperforms other methods, which improves by 2 percentage points to $94\%$ comparing with STGAN. It is also seen in Fig \ref{fig:six} that the reconstructed images yielded by our method is more consistent with source images in each aspect (background, details, etc) than other models.

\begin{table}
	\begin{center}
		\begin{tabular}{ccccc}
			\hline
			Method& StarGAN&AttGAN&STGAN&ClsGAN \\ \hline
			F,S&7.9/0.56&7.1/0.8&6.1/0.92&6.09/0.94
			\\ \hline	
		\end{tabular}
	\end{center}
	\caption{Quality and reconstruction performance of the comparison methods on facial attribute editing task.}
	\label{Tab1}
\end{table}

\begin{figure}
	\centering
	\includegraphics[width=0.7\textwidth]{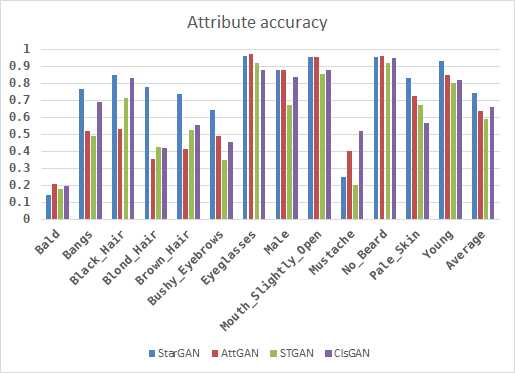}
	\caption{The attribute accuracy about StarGAN, AttGAN, STGAN and ClsGAN.}
	\label{fig:eight}
\end{figure}

As for the classification accuracy, the training set of CelebA dataset is adopted to train a classifier for $13$ attributes and attain the average accuracy of 93.83\% in the test set. Then the pre-trained classifier is used to test the transfer accuracy of different models between 2000 synthetic images. In order to compare the editing ability of each model, the conversion rate of 13 attributes is listed in the form of a bar chart. As can be seen in Figure \ref{fig:eight}, the performance of StarGAN is outstanding in hair color transfer. A possible reason is that the hair color is more prone to be affected by source attribute information owing to ease of color conversion, while the StarGAN avoids the additional introduction of source attribute information by using no skip-connection layers. Comparing with STGAN, ClsGAN attains significant gain in attribute Bangs (0.689), Black\_hair (0.8325) and Mustache (0.523), which profits from the Atta-cls method. The transfer accuracy of other attributes is also relatively remarkable comparing with AttGAN and STGAN and has average transfer 0.66. Although the accuracy is slightly lower than StarGAN(average 0.74), image quality(FID 6.09) is much better than StarGAN(FID 0.79). It implicates that our attribute adversarial classifier which implements attribute generation in an adversarial way is effective to enhance the transfer accuracy. The accuracy of attribute Pale is relatively poor in numerical terms, which is likely that the pre-trained classifier can't recognize the image with a lighter conversion effect. However, it is acceptable to the editing effect visually (see Fig \ref{fig:five}).

\begin{figure*}
	\centering
	\subfigure[Season transfer]{
		\includegraphics[height=0.25\textheight,width=0.48\textwidth]{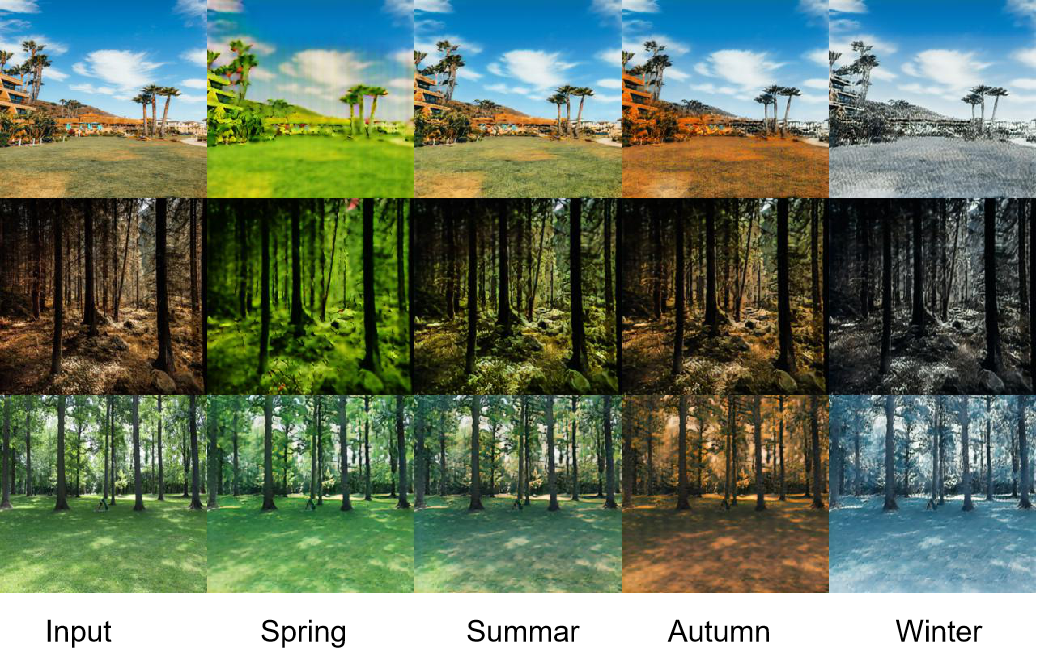}}
	\hspace{0.01in}
	\subfigure[Artistic transfer]{
		\includegraphics[height=0.25\textheight,width=0.48\textwidth]{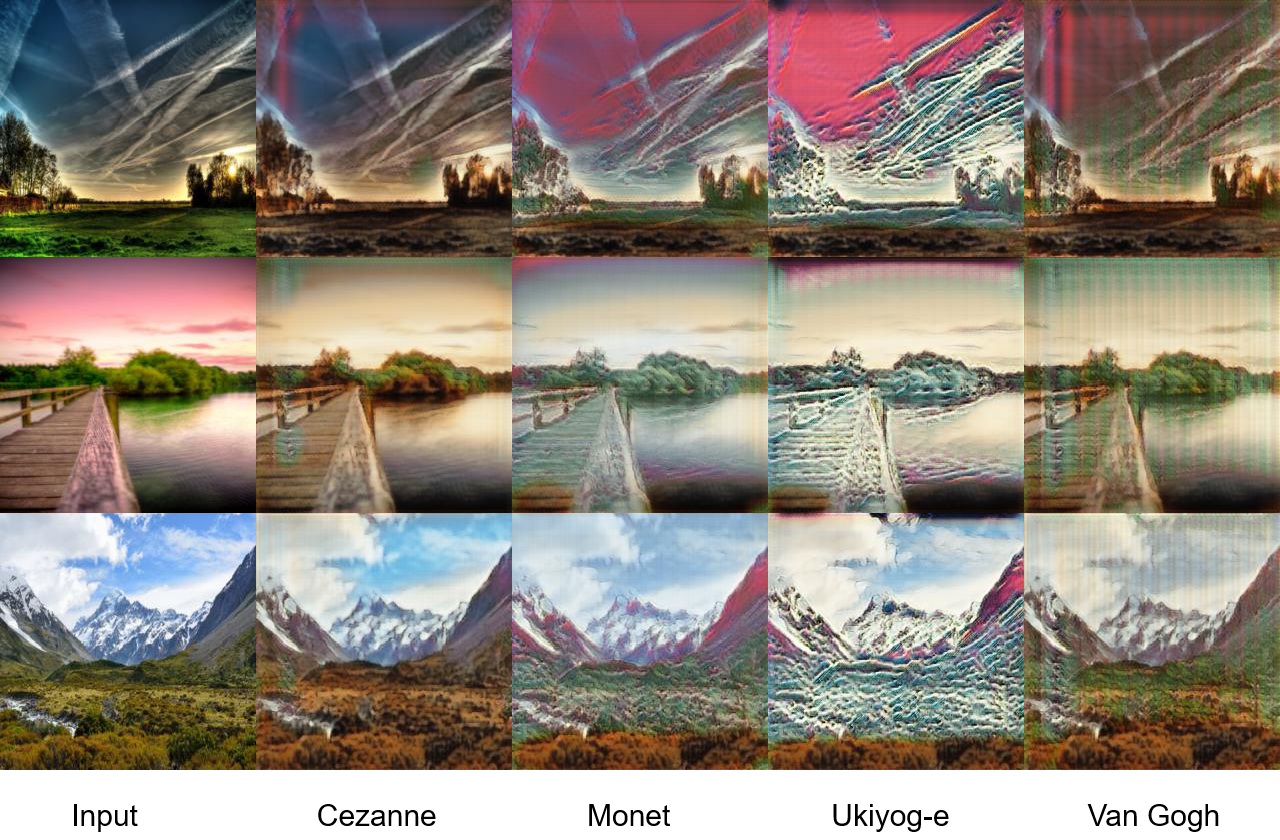}}
	\caption{The $256\times 256$ transfer images about season dataset and painting dataset. Please zoom in for better observation.}
	\label{fig:nine}	
\end{figure*}

\subsection{Seasons and artistic styles transfer}
Since the objective of style transfer is the same with attribute editing to some extent, ClsGAN is also implemented to realize the style transformation task. The method is employed on a season dataset and a painting dataset. Seasonal images are from the Unplash website, where the number of images of different seasons is: spring ($29343$), summer ($23395$), fall($7630$) and winter($13433$). The painting images mainly come from the wikiart website, and ClsGAN achieves the mutual transformation between four styles and photographs. The number of images is Monet: $1050$, Cezanne: $582$, VanGogh: $1931$, Ukiyo-e: $1372$, Photograph: $4674$. The photographs are downloaded from Flickr and use landscape labels, and are all resized as $256\times 256$. 

It can be seen from the Fig \ref{fig:nine} that the result about the artistic and season transfer is acceptable, but there are some artifacts in some synthetic images (the second image in the first row of (a), the last column of (b)). One possible reason is that it is limited when editing the large range texture and color using a single model. On the other hand, the attribute editing model may not be able to balance the effect between image quality and attribute transformation, because it needs to pay more attention to a lot of texture information. However, ClsGAN is a potential model which is still deserved to further explore and extent. 

\subsection{Ablation Study}

\begin{figure*}
	\centering	
	\includegraphics[width=1\textwidth]{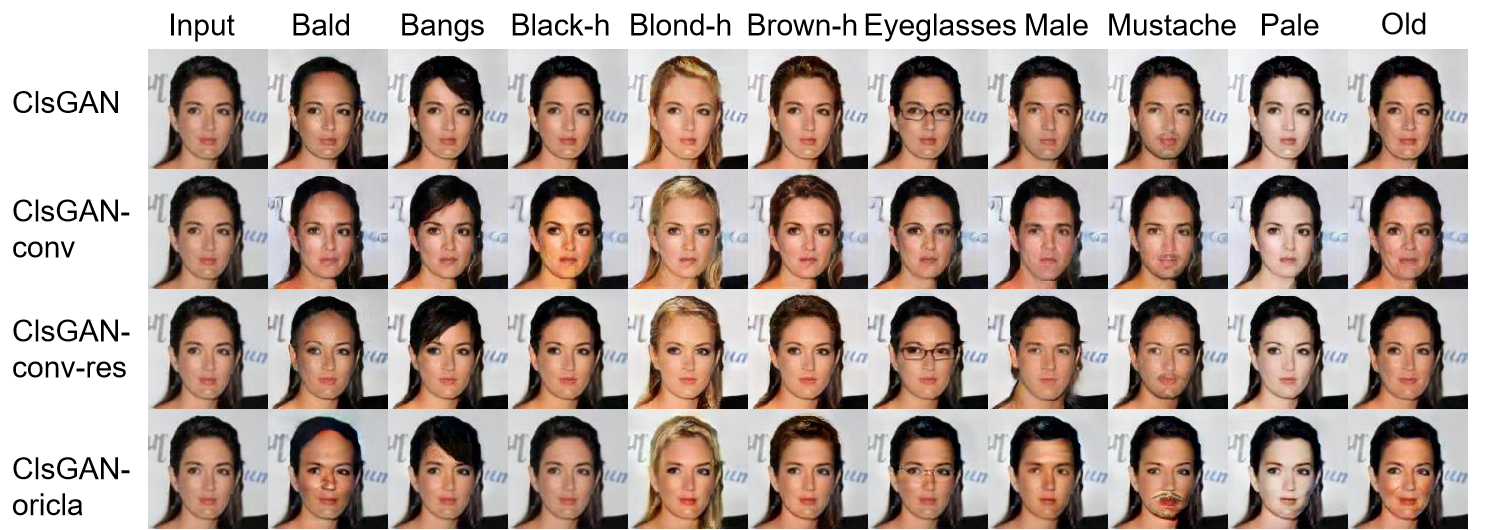}
	\caption{Face transfer results on four different combinations. Please zoom in for better observation.}
	\label{fig:ten}
\end{figure*}

In this part, the roles of Tr-resnet and attribute adversarial classifier (Atta-cls) are investigated. Concretely, four different combinations is considered: (i) ClsGAN: the original model; (ii)ClsGAN-conv: substituting Tr-resnet with the convolution network in the decoder; (iii)ClsGAN-conv-res: adopting the residual technique to learn the convolution in ClsGAN-conv; (iv)ClsGAN-oricla: adopting the original classifier which is used in StarGAN, AttGAN and STGAN instead of the adversarial classifier. Fig \ref{fig:ten} shows the results of different variants.

\textbf{Tr-resnet vs its variants}
In Fig\ref{fig:ten} (row(1),row(2),row(3)), the results about Tr-resnet and its variants is showed. It can be seen that the images outperform the other combinations using Tr-resnet. Compared with ClsGAN, the results of ClsGAN-conv is undesired in some attributes, e.g., Eyeglasses, Old. The situation implies that it is insufficient to attain the necessary information from attributes label only using convolution operation. The images generated by ClsGAN-conv-res are relatively acceptable but the global editing contains the unnecessary changes(such as Male editing alters the hairstyle in the third row). 
As can be seen, the transfer images of ClsGAN are more accurate to all attributes and the quality has better performance.

\textbf{Atta-cls vs original classifier}
Compared with Atta-cls, the results(see row(4)) of using the original classifier look more unrealistic accompanied with artifacts and blurriness. One available reason is that only using the original classifier is likely to affect the gain of the necessary information from original images and attributes label, so as lower  performance of photo-realistic and attribute accuracy. In this paper, the adversarial technique is adapted to motivate the classifier to learn detailed attribute information, thus it improves the quality of images by optimizing the generator. From row(1) of Fig \ref{fig:ten}, it can see that the results of our ClsGAN method look more photo-realistic and nature compared with the original classifier method.

\section{Conclusion}
In this paper, the work first analyzes the constraint problem between image attribute transfer and quality
about attribute editing and proposes the ClsGAN model by incorporating the upper convolution residual network(Tr-resnet) and attribute adversarial classifier(Atta-cls). About Tr-resnet, the upper convolution structure is applied with residual technique to select the desired information. Specially, Atta-cls is presented to enhance the attribute transfer accuracy of the image, which is inspired by the spirit of generated adversarial network. The attribute adversarial classifier can selectively find necessary information about attributes and then optimize the generator by an adversarial way. At the same time, an approximation between the source label and attribute feature vector(which is generated by style encoder) is made to meet the requirement of label continuity. Experiments and ablation studies both demonstrate the great effectiveness of ClsGAN in attribute editing.

\section*{Acknowledgement}
This work was primarily supported by National Key R\&D Program of China (NO.2018YFC1604000) and Foundation Research Funds for the Central Universities (Program No.2662017JC049) and State Scholarship Fund (NO.261606765054).

\bibliographystyle{unsrt}
\bibliography{references}  %%% Remove comment to use the external .bib file (using bibtex).


\end{document}